\documentclass[11pt]{article}

\usepackage[preprint]{acl}

\usepackage{times}
\usepackage{latexsym}

  \usepackage{amsmath,amssymb}
  \usepackage{algorithm}
  \usepackage{algorithmic} % 若用 algpseudocode 可替换为
\usepackage[T1]{fontenc}
\usepackage[utf8]{inputenc}

\usepackage{microtype}

\usepackage{inconsolata}

\usepackage{xcolor}
\usepackage{graphicx}
\usepackage{url}
\usepackage{hyperref}

\usepackage{graphicx}
\usepackage{amsmath}
\usepackage{amssymb}
\usepackage{booktabs}

\usepackage{mathtools}
\usepackage[nolist]{acronym} % Basic package for the definition of special characters in sign language

\usepackage{bm}

\usepackage{caption}
\usepackage[nolist]{acronym}
\begin{acronym}[PHOENIX14\textbf{T} ]

\acrodefplural{rnn}[RNNs]{Recurrent Neural Networks}
\acrodefplural{cnn}[CNNs]{Convolutional Neural Networks}
\acrodefplural{hmm}[HMMs]{Hidden Markov Models}
\acrodefplural{gru}[GRUs]{Gated Recurrent Units}
\acrodefplural{crf}[CRFs]{Conditional Random Fields}
\acrodefplural{gan}[GANs]{Generative Adversarial Networks}
\acrodefplural{gpu}[GPUs]{Graphic Processing Units}

\acrodefplural{mdn}[MDNs]{Mixture Density Networks}

\acro{aslt}[ASLT]{American Sign Language Translation}
\acro{aslp}[ASLP]{American Sign Language Production}
\acro{bsl}[BSL]{British Sign Language}
\acro{bleu}[BLEU]{Bilingual Evaluation Understudy}
\acro{blstm}[BLSTM]{Bidirectional Long Short-Term Memory}
\acro{bslcpt}[BSLCP\textbf{T}]{BSL Corpus \textbf{T}}
\acro{cnn}[CNN]{Convolutional Neural Network}
\acro{crf}[CRF]{Conditional Random Field}
\acro{cslr}[CSLR]{Continuous Sign Language Recognition}
\acro{ctc}[CTC]{Connectionist Temporal Classification}
\acro{c4a}[C4A]{Content4All}
\acro{dl}[DL]{Deep Learning}
\acro{dgs}[DGS]{German Sign Language - Deutsche Gebärdensprache}
\acro{dsgs}[DSGS]{Swiss German Sign Language - Deutschschweizer Geb\"ardensprache}
\acro{dtw}[DTW]{Dynamic Time Warping}
\acro{fc}[FC]{Fully Connected}
\acro{ff}[FF]{Feed Forward}
\acro{gan}[GAN]{Generative Adversarial Network}
\acro{gpu}[GPU]{Graphics Processing Unit}
\acro{gru}[GRU]{Gated Recurrent Unit}
\acro{gslp}[GSLP]{German Sign Language Production}
\acro{gtpt}[G2PT]{Gloss to Pose Transformer}
\acro{hmm}[HMM]{Hidden Markov Model}
\acro{hpe}[HPE]{Hand Pose Enhancer}
\acro{isl}[ISL]{Irish Sign Language}
\acro{lstm}[LSTM]{Long Short-Term Memory}
\acro{mha}[MHA]{Multi-Headed Attention}
\acro{mtc}[MTC]{Monocular Total Capture}
\acro{mse}[MSE]{Mean Squared Error}
\acro{mslp}[MSLP]{Multilingual Sign Language Production}
\acro{mdn}[MDN]{Mixture Density Network}
\acro{mdgs}[mDGS]{meineDGS}
\acro{mdgsv}[mDGS-V]{meineDGS-Variants}
\acro{mdgst}[mDGS\textbf{T}]{meineDGS\textbf{T}}
\acro{mdgsth}[mDGS\textbf{T}-\textbf{H}]{meineDGS\textbf{T}-\textbf{HARD}}
\acro{mdgste}[mDGS\textbf{T}-\textbf{E}]{meineDGS\textbf{T}-\textbf{EASY}}

\acro{nmt}[NMT]{Neural Machine Translation}
\acro{nlp}[NLP]{Natural Language Processing}
\acro{ph12}[PHOENIX12]{RWTH-PHOENIX-Weather-2012}
\acro{ph14}[PHOENIX14]{RWTH-PHOENIX-Weather-2014}
\acro{ph14t}[PHOENIX14\textbf{T}]{RWTH-PHOENIX-Weather-2014\textbf{T}}
\acro{pof}[POF]{Part Orientation Field}
\acro{paf}[PAF]{Part Affinity Field}
\acro{pttt}[P2TT]{Pose to Text Transformer}
\acro{relu}[RELU]{Rectified Linear Units}
\acro{rnn}[RNN]{Recurrent Neural Network}
\acro{rouge}[ROUGE]{Recall-Oriented Understudy for Gisting Evaluation}
\acro{sgd}[SGD]{Stochastic Gradient Descent}
\acro{sla}[SLA]{Sign Language Assessment}
\acro{slr}[SLR]{Sign Language Recognition}
\acro{slt}[SLT]{Sign Language Translation}
\acro{slp}[SLP]{Sign Language Production}
\acro{smt}[SMT]{Statistical Machine Translation}

\acro{ttgt}[T2GT]{Text to Gloss Transformer}
\acro{ttpt}[T2PT]{Text to Pose Transformer}
\acro{ttp}[T2P]{Text to Pose}
\acro{ttgtp}[T2G2P]{Text to Gloss to Pose}
\acro{tts}[TTS]{Text to Speech}
\acro{wer}[WER]{Word Error Rate}

\end{acronym}
\usepackage{dblfloatfix}

\usepackage{multirow}

\newcommand{\methodName}{\textsc{SignGAN}}
\usepackage{enumitem,xcolor}
\newlist{coloritemize}{itemize}{1}
\setlist[coloritemize]{label=\textcolor{itemizecolor}{\textbullet},font=\bfseries\color{itemizecolor}}
\usepackage{accents}
\usepackage[accsupp]{axessibility}  % Improves PDF readability for those with disabilities.
\usepackage[title]{appendix}

\title{Stable Signer: Hierarchical Sign Language Generative Model}
\author{Sen Fang$^{1}$, Yalin Feng$^{2}$\thanks{Equal Contribution.}, Hongbin Zhong$^{3}$, Yanxin Zhang$^{4}$, Dimitris N. Metaxas$^{1}$ \\
   $^{1}$Rutgers University, $^{2}$Nanyang Technological University, $^{3}$Georgia Institute of Technology, \\$^{4}$University of Wisconsin-Madison\\
   \url{https://stablesigner.github.io/}
   }

\begin{document}
\maketitle

\begin{abstract}

%手语制作是将输入的复杂文本转换成真实视频的过程，先前的工作大部分集中在Text2Gloss，Gloss2Pose，Pose2Vid等环节上（Gloss是表达特定手势的文本词汇），也有一些集中在Prompt2Gloss和Text2Avatar环节上。但是这个领域进展缓慢，由于文本转换、姿势生成、姿势渲染成真人视频这些环节的不准确，误差逐渐累积。因此在这篇文章中，我们抛弃传统的冗余结构，精简优化任务目标，设计了一种名为Stable Signer的新手语Generative Model。它将手语制作任务重新定义为只包括文本理解（Prompt2Gloss、Text2Gloss）和Pose2Vid的分层生成的端对端任务，通过我们提出的新多语种语言理解层称为SLUL来执行文本理解，通过名为SLP-MoE的手语渲染生成专家块 端对端生成高质量、多风格的手语视频。SLUL使用新开发的多语种Semantic-Aware Gloss Masking Loss (SAGM Loss) 训练。性能比目前最先进的生成方法提高了27%，这在SLP领域里是巨大的增幅。
Sign Language Production (SLP) is the process of converting the complex input text into a real video. Most previous works focused on the Text2Gloss, Gloss2Pose, Pose2Vid stages\footnote{Gloss represents the specific gesture text words.}, and some concentrated on Prompt2Gloss and Text2Avatar stages. However, this field has made slow progress due to the inaccuracy of text conversion, pose generation, and the rendering of poses into real human videos in these stages, resulting in gradually accumulating errors. Therefore, in this paper, we streamline the traditional redundant structure, simplify and optimize the task objective, and design a new sign language generative model called \textbf{Stable Signer}. It redefines the SLP task as a hierarchical generation end-to-end task that only includes text understanding (Prompt2Gloss, Text2Gloss) and Pose2Vid, and executes text understanding through our proposed new \textbf{S}ign \textbf{L}anguage \textbf{U}nderstanding \textbf{L}inker called \textbf{SLUL}, and generates hand gestures through the named \textbf{SLP-MoE} hand gesture rendering expert block to end-to-end generate high-quality and multi-style sign language videos. SLUL is trained using the newly developed \textbf{S}emantic-\textbf{A}ware \textbf{G}loss \textbf{M}asking Loss (\textbf{SAGM Loss}). Its performance has improved by 48.6\% compared to the current SOTA generation methods,
%Project Page at \url{https://stablesigner.github.io/}.
which is a significant increase in the SLP field. %More demo can be obtained at \href{https://anonymoussubmissionurl.github.io/Stable-Signer/}{anonymous url}.
\end{abstract}
\section{Introduction}

%在基于深度学习的手语领域，手语识别首先发展并流行起来，然后大约在2015-2020年之后，手语制作也逐渐的流行起来。不过相比于成熟而又流行的手语识别、手语理解工作，手语制作相对于落后。这是有多个原因造成的：（1）不同数据集姿势处理方法不同，而且精准度有差异，这使得姿势的学习相对困难。（2）不同的生成或评估模型是专用某种姿势格式的，导致比较迁移相对困难。（3）手语制作的环节相比识别要更多，环节误差累积导致效果不容易提升。
In the field of sign language based on deep learning, Sign Language Recognition (SLR \cite{hu2023continuous,skeleton_aware_slr,slt-how2sign-wicv2023}) first developed and became popular, and then around 2015-2020, Sign Language Production (SLP \cite{SLP_Mixed,SLP_ProgTransf,Fang_2025_ICCV}) also gradually became popular. However, compared to the mature and popular SLR and understanding tasks, \textit{sign language production is relatively lagging behind}. This is caused by several reasons: \textcolor{purple}{\textbf{(1)}} The processing methods for different datasets vary, and there are differences in accuracy, which makes the learning of postures relatively difficult. \textcolor{purple}{\textbf{(2)}} Different generation or evaluation models are specific to certain posture formats, making the comparison and transfer relatively difficult. \textcolor{purple}{\textbf{(3)}} The multi-stage SLP pipeline suffers from \textit{error accumulation across stages}, making performance improvements challenging.

\begin{figure}[t]
  \centering
   \includegraphics[width=0.99\linewidth]{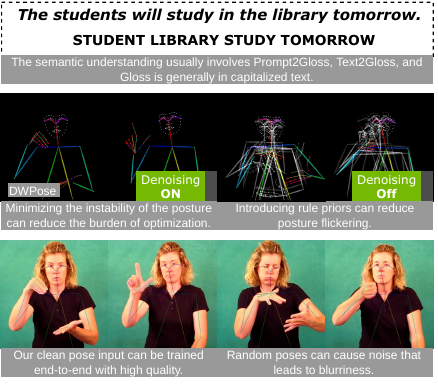}
   \vspace{-20pt}
   \caption{\textbf{The current SLP paths and their drawbacks:} SLP contains that the prompt/complex text gradually transforms into a pose video, and then it goes through the process of being converted into a real person video. However, this process is too complex to involves a lot of errors, which accumulate. Therefore, we plan to reduce the redundancy of unnecessary intermediate steps (as shown in the figure, same ``Gloss'' has poses that do not correspond in time), make the initial and final steps more closely connected, and then achieve end-to-end hierarchical model learning.}
   \label{fig:cover}
   \vspace{-12pt}
\end{figure}

%为了解决这些问题，参考手语制作的一些经典工作，我们意识到基于深度学习的手语制作的环节存在冗余：SLP任务的输入输出是文本和最终的高质量视频，对于其中的姿势查找方法，其实可以使用规则作为先验，因为它只需要正确，即gloss和文本只需要对应一个确定的姿势或者手势。手语识别无法引入规则先验是因为存在多种相似姿势输入的可能，而手语制作只要精确的姿势输出，手语制作核心问题其实是gloss的输入需要准确，因此压力在语言理解部分。
To address these issues, drawing on some of the similar work \cite{shazeer2017outrageouslylargeneuralnetworks,riquelme2021scalingvisionsparsemixture} in Sign Language Production (SLP), we realized that there is \textit{redundancy in the process of SLP based on deep learning}: the input and output of the SLP task are text and the final high-quality video. For pose video, \textbf{rules can be used as a prior} \cite{Stoll2020text2sign}, because the gloss and the text only need to correspond to a specific gesture or movement. The SLR cannot introduce rule priors because there are multiple possible inputs of similar gestures, while in SLP, as long as the precise gesture output is required, \textit{the core problem of SLP is actually that the input of the gloss needs to be accurate}, so the pressure lies in the language understanding \cite{NEURIPS2022_ec795aea} part.

%如果我们引入基于规则的方法作为先验，我们可以减小整个姿势环节的误差，甚至可以确保这个环节几乎不再出问题，真正的
%而且更严重的是，近期的手语制作方法其实没有考虑到它们需求是和SLR、SLT不同的。如Fig1所示，对于一个确定的姿势，有多种表达，如果我们强行学习姿势的生成，就容易陷入到平均姿势（次优状态）。如果我们的训练目标或者数据没有时刻精确对齐，姿势越多反而学习越不准确或者困难。因此，端对端的动机和现存在的两个问题促使我们将这个冗余的部分去掉。然后我们就可以将真正需要的语言理解、Pose2Vid部分进行研究，提升。真正大规模端对端的学习真正需要的部分。
And what's more serious is that the SLP methods actually did not take into account that their requirements are different from those of SLR and SLT. As shown in Fig \ref{fig:cover}, for a specific posture, there are \textit{multiple expressions}. If we forcibly learn the generation of the posture, we are likely to fall into the \textit{average posture (suboptimal state)}. If our training goals or data are not always precisely aligned, the more postures there are, the less accurate or difficult it is to learn at same time. Therefore, the motivation of end-to-end and the two existing problems prompt us to \textbf{streamline this redundant part} \cite{saunders2020progressivetransformersendtoendsign, saunders2021continuous3dmultichannelsign}. Then we can focus on researching and improving the truly necessary parts of language understanding \cite{NEURIPS2022_ec795aea} and Pose2Vid.
%, which are the truly large-scale end-to-end learning requires.
%Therefore, the current SLP methods do not actually take into account that the real requirements are different from SLR. As shown in Fig. \ref{fig:cover}, if we introduce a rule-based method as a prior, we can reduce the error in the entire gesture process, and even ensure that this process rarely has problems, truly eliminating this redundant part \cite{saunders2020progressivetransformersendtoendsign, saunders2021continuous3dmultichannelsign}. Then we can focus on researching and improving the truly necessary language understanding \cite{NEURIPS2022_ec795aea} and Pose2Vid parts. When we eliminate this redundant part, we can truly conduct large-scale end-to-end learning of the truly necessary parts.

%既然我们可以前所未有的端对端的学习SLP，那么就要考虑文本理解和姿势条件视频生成将会面临的挑战：（1）传统的SLP使用Gloss或者Text作为输入，为了准确，他们也会使用翻译模型将Prompt或复杂Text翻译成Gloss。这其实相当于去掉冗余的姿势的随机，只是没有我们彻底，因此不够简洁。为了减小冗余反而可能引入了更多误差。（2）在多语种处理时，以前的很多方法为了处理异构的手语数据，多次重复训练模型或者专门创建方法，这不必要的增加了任务的复杂性。（3）对于Pose2Video部分，先前的方法由于没有端对端的训练，很难将复杂的文本信息用于辅助，或者将姿势输入进行更加紧密的结合。这是很难提升最终视频效果的主要原因。
Since we can now learn SLP in an rarely end-to-end manner, we need to consider the challenges that text understanding and pose-conditioned video generation will face: \textcolor{purple}{\textbf{(1)}} Traditional SLP uses Gloss or Text as input \cite{inproceedings}. To ensure accuracy, they also use translation models to translate Prompts or complex Text into Gloss, but it is not as thorough as ours.
%This is actually equivalent to removing the redundant random of the pose, but it is not as thorough as ours, thus not being as concise. To reduce redundancy, it may instead introduce more errors. 
\textcolor{purple}{\textbf{(2)}} When dealing with processing, many previous methods, in order to handle heterogeneous sign language data, repeatedly train the model or specially create methods, which unnecessarily increases the complexity of the task. \textcolor{purple}{\textbf{(3)}} For the Pose2Video \cite{chan2019everybodydance,saunders2020everybodysignnowtranslating} part, previous methods, due to the lack of end-to-end training, have difficulty using complex textual information for assistance or combining the pose input of Pose2Vid model more closely. This is the main reason why it is difficult to improve the final video effect.

%为了解决这些挑战，我们统一将核心设置为基于规则的可学习混合式Gloss2Pose，然后设计一个手语理解层叫SLUL，它可以将多语种Prompt或者Text统一翻译成Gloss，它使用我们开发的Prompt Mask Loss进行训练，这个新Loss用来解决不同语种输入，以及防止Gloss冲突。然后我们开发SLP-MoE模块，将不同语种输入的语义信息输入到Pose2Vid模型的输入当中，用于端对端训练。之前的手语Pose2Vid模型由于制作链条过长存在冗余，无法端对端训练，我们这个方法可以将以前方法不曾利用的语义信息，更多的姿势信息都利用起来，使得生成的视频质量更高，语义表达更好。
To address these challenges, we uniformly set the core as a \textit{rule-based learnable hybrid} Gloss2Pose\footnote{We introduce prior knowledge not by completely giving up learning, but by ingeniously transforming the learning objective into achieving the optimal posture and optimizing the smoothness of transitions.}. Then, we designed a sign language understanding linker called \textbf{SLUL}, which can uniformly translate prompts or texts into Gloss. It is trained using the Prompt Mask Loss developed by us. This new Loss is used to handle different language inputs and prevent Gloss conflicts. Subsequently, we developed the \textbf{SLP-MoE module}, which inputs the semantic information of different language inputs into the Pose2Vid model's input for end-to-end training. 
%The previous sign language Pose2Vid model had redundant production chains due to a long production chain, making it impossible for end-to-end training \cite{saunders2020progressivetransformersendtoendsign, saunders2021continuous3dmultichannelsign}. 
Our method can utilize the semantic information and more posture information that were not utilized by previous methods, resulting in \textit{higher video quality and better semantic expression}.

%综上所述，我们的贡献可以总结为：
%一个用于统一多语种理解的手语理解模块SLUL，可以将复杂的提示词或者文本手语信息精确转换成不同手语所需的Gloss。
%多语种手语视频生成MoEs，可以接收SLUL输出信息，结合语义将Gloss快速准确的生成精准的Pose Video，然后进而渲染成多样高质量手语视频。
% 一个UPM Loss，第一次可以在2D手语视频上实现端对端的训练，并且实验显示我们比现有最新工作效果提升48.6%，这在SLP领域是巨大的提高。
In summary, our contributions can be summarized as follows:
\begin{itemize}
    \item A \textbf{Sign Language Understanding Linker (SLUL)} for unified understanding, which can accurately convert complex prompts or textual sign language information into Glosses required for different sign languages.
    \item \textbf{Sign Language Production Mixture-of-Experts (SLP-MoEs)}, which can receive output from SLUL and rapidly and accurately generate precise pose videos from Glosses, subsequently rendering them into diverse high-quality sign language videos.
    \item A \textbf{Semantic-Aware Gloss Masking (SAGM) Loss}, which enables end-to-end training on 2D sign language videos. Experiments demonstrate a \textit{48.6\% improvement} over the state-of-the-art, which represents a substantial advancement in the SLP domain.
\end{itemize}

\begin{figure*}[t]
  \centering
   \includegraphics[width=0.99\linewidth]{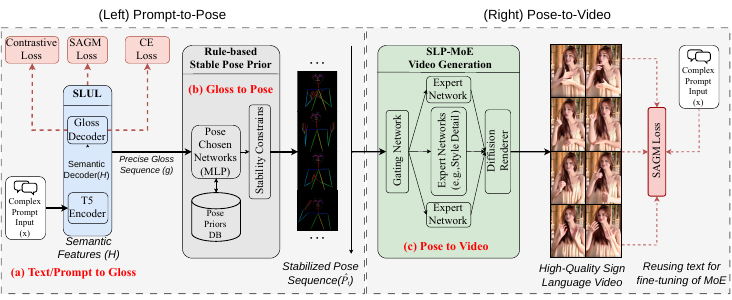}
   \vspace{-6pt}
   
\caption{\textbf{Overview of the Prompt2Pose and Pose2Video pipeline:} 
\textcolor{purple}{\textbf{(a) Text/Prompt to Gloss:}} The SLUL module uses a T5 encoder to process complex prompts and generate precise gloss sequences, trained with SLUL Loss, SAGM Loss, KL divergence, and contrastive loss. 
\textcolor{purple}{\textbf{(b) Gloss to Pose:}} A rule-based stable pose prior database provides candidate poses, which are selected by the Pose Chosen Networks (MLP) guided by semantic features and trained to ensure natural and smooth pose transitions.
%The Gating Network and Expert Networks in SLP-MoE determine the optimal pose selection with stability constraints applied. 
\textcolor{purple}{\textbf{(c) Pose to Video:}} The stabilized pose sequence is fed into a diffusion renderer to generate high-quality sign language videos in multiple styles, with the semantic features reused for fine-tuning the MoE module.}
   \label{fig:method}
   \vspace{-4pt}
\end{figure*}

\section{Methodology}

\subsection{Data Construction}

\textbf{For the sign language}, we use a dataset called Prompt2Sign \cite{Fang_2025_ICCV}. The videos in this dataset are sourced from existing mainstream datasets and online sign language videos, and they are segmented into tens of thousands of clips. It has LLM-generated prompts, texts, and comprehensive glosses. It also provides a unified processing procedure to facilitate our comparison with existing methods. \textbf{For pose condition control}, we first use a video dataset called OpenVidHD\footnote{\url{https://github.com/NJU-PCALab/OpenVid-1M}} \cite{nan2025openvid1mlargescalehighqualitydataset} for training to improve video quality. Then, we use sign language videos to enhance the posture guidance ability.

\subsection{Linking Semantics with Sign Language Understanding}
\label{subsec:slul}
We cast prompt/text $\mathbf{x}$ (tagged by language identifier $\ell$) into a sequence-to-sequence task that predicts a gloss sequence $\mathbf{g}=(g_1,\ldots,g_T)$. A T5 encoder produces contextual states $H=f_{\theta_E}([\ell;\mathbf{x}])\in\mathbb{R}^{L\times d}$, where $L$ is the input sequence length and $d$ is the hidden dimension; the decoder autoregressively generates gloss tokens with maximum-likelihood training
\[
  \mathcal{L}_{\text{SLUL}}=-\sum_{t=1}^T \log p_\theta(g_t \mid g_{<t},H).
\]
This base term aligns prompts with gloss semantics while sharing encoder parameters.
%across languages to enable cross-lingual transfer and zero-/few-shot glossing.

\noindent\textbf{Semantic Aware Gloss Masking Loss (SAGM Loss).}
To reduce gloss ambiguity and force semantic reconstruction, we randomly mask gloss tokens (mask rate $\rho$), forming $\tilde{\mathbf{g}}$ and requiring the model to infer the masked entries. The masked objective is
\[
  \mathcal{L}_{\text{SAGM}}=-\sum_t \mathbf{1}[u_t\le\rho]\,
    \log p_\theta(g_t \mid \tilde{\mathbf{g}},H),
\]
where $u_t\sim\mathcal{U}(0,1)$. We further enforce consistency between masked/unmasked posteriors
\[
  \mathcal{L}_{\text{KL}}=\mathrm{KL}\!\big(p_\theta(\cdot \mid
\tilde{\mathbf{g}},H)\,\|\,p_\theta(\cdot \mid \mathbf{g},H)\big),
\]
so the decoder cannot drift when surface cues vanish. Intuitively, SAGM behaves as a semantic denoiser that combats noisy or rare gloss forms.

\noindent\textbf{Stable SLUL and Contrastive Linking.}
We prepend $\ell$ to inputs and couple prompt/gloss embeddings via contrastive alignment:
\[
  \mathcal{L}_{\text{con}}
  =-\log \frac{\exp(\langle \bar{h}_x,\bar{h}_g\rangle/\tau)}
  {\sum_{g'} \exp(\langle \bar{h}_x,\bar{h}_{g'}\rangle/\tau)},
\]
with mean-pooled embeddings $\bar{h}_x=\mathrm{Pool}(H)$, $\bar{h}_g=\mathrm{Pool}(f_{\theta_D}(\mathbf{g}))$ from encoder and decoder outputs respectively, and temperature $\tau$. Following standard practice, negative samples $g'$ are drawn from the same batch. The semantic objective becomes
\begin{align*}
  \mathcal{L}_{\text{SLUL+SAGM}}
  ={}&\mathcal{L}_{\text{SLUL}}
  +\lambda_{\text{SAGM}}\mathcal{L}_{\text{SAGM}} \\
  &+\lambda_{\text{KL}}\mathcal{L}_{\text{KL}}
  +\lambda_{\text{con}}\mathcal{L}_{\text{con}}.
\end{align*}
This stack enforces (i) faithful prompt-to-gloss mapping, (ii) robustness to masked cues, and (iii) cross-lingual embedding agreement.

\paragraph{Pseudo-code (semantic stage).} This semantic stage equips SLUL with robustness to noisy or under-specified prompts: masking forces the decoder to rely on encoder semantics, KL keeps masked/unmasked posteriors aligned, and contrastive linking ties prompt and gloss embeddings across languages. However, accurate gloss alone does not guarantee stable video; pose retrieval and temporal stability are critical. We therefore turn to a gated mixture-of-experts to select and blend pose priors before rendering.

\begin{algorithm}[t]
\small
\caption{SLUL + SAGM forward}
\begin{algorithmic}[1]
  \STATE $H \gets f_{\theta_E}([\ell;\mathbf{x}])$
  \STATE Sample masks $m_t \sim \mathrm{Bernoulli}(\rho)$
  \STATE Set $\tilde{g}_t \gets \text{\texttt{[MASK]}}$ if $m_t = 1$ else $g_t$
  \STATE Compute $p_\theta(\cdot \mid \tilde{\mathbf{g}},H)$ and $p_\theta(\cdot \mid \mathbf{g},H)$
  \STATE $\mathcal{L}_{\text{SLUL+SAGM}} \gets \mathcal{L}_{\text{SLUL}} + \lambda_{\text{SAGM}}\mathcal{L}_{\text{SAGM}} + \lambda_{\text{KL}}\mathcal{L}_{\text{KL}} + \lambda_{\text{con}}\mathcal{L}_{\text{con}}$
\end{algorithmic}
\end{algorithm}

\subsection{SLP-MoE for Stable Pose2Video Production}
\label{subsec:slp_moe}
Given predicted gloss $\mathbf{g}$ and semantic features $H$ from SLUL, we select pose priors from a rule-based database with a gated mixture-of-experts (MoE), then stabilize keypoints before rendering.

\noindent\textbf{SLP-MoE: Gated Pose Experts.}
A gloss-conditioned query $q=\mathrm{Pool}(H)$ produces gates over $K$ experts:
\begin{align*}
  w_k&=\frac{\exp(q^\top W_k)}{\sum_{j=1}^K \exp(q^\top W_j)},\\
  \mathbf{p}_{\text{pose}}&=\sum_{k=1}^K w_k\,\phi_k(\mathbf{g}),
\end{align*}
where each expert $\phi_k$ retrieves a candidate pose sequence from the database based on gloss $\mathbf{g}$, and $\mathbf{p}_{\text{pose}}$ represents the weighted blend of these retrieved sequences. During training with ground truth pose sequence index $y$, we optimize
\begin{align*}
  \mathcal{L}_{\text{MoE}}&=-\log\!\sum_k w_k \mathbf{1}[k=y],\\
  \mathcal{L}_{\text{ent}}&=-\sum_k w_k \log w_k,
\end{align*}
encouraging correct selection and avoiding gate collapse. At inference, we use the expert with highest gate weight.

\noindent\textbf{Stability-Constrained Pose Blending.}
We refine the blended pose sequence $\mathbf{p}_{\text{pose}}$ into stabilized keypoints $\hat{P}_t \in \mathbb{R}^{J \times 2}$ for $J$ body keypoints at frame $t$, extracting hand keypoint subset $\hat{H}_t$. We penalize jerk and preserve hand fidelity:
\begin{align*}
  \mathcal{L}_{\text{smooth}}&=\sum_t \|\hat{P}_t-2\hat{P}_{t-1}+\hat{P}_{t-2}\|_2^2,\\
  \mathcal{L}_{\text{hand}}&=\sum_t \|\hat{H}_t-H_t^\star\|_2^2,
\end{align*}
where $H_t^\star$ denotes ground truth hand keypoints. We optionally damp micro-flicker via $\mathcal{L}_{\text{vel}}=\sum_t \|\hat{P}_t-\hat{P}_{t-1}\|_2^2$. These terms directly target temporal jitter and hand intelligibility.

\noindent \textbf{Pose-to-Video Conditioning.}
The stabilized pose sequence $\{\hat{P}_t\}$ drives a diffusion renderer (ControlNeXt/Wan) as control signals. The hierarchical objective coupling semantics, pose selection, and stability is
\begin{equation*}
\begin{aligned}
  \mathcal{L}
  ={}&\mathcal{L}_{\text{SLUL+SAGM}}
  +\lambda_{\text{MoE}}\mathcal{L}_{\text{MoE}}
  +\lambda_{\text{ent}}\mathcal{L}_{\text{ent}} \\
  &+\lambda_{\text{smooth}}\mathcal{L}_{\text{smooth}}
  +\lambda_{\text{hand}}\mathcal{L}_{\text{hand}}
  +\lambda_{\text{vel}}\mathcal{L}_{\text{vel}}.
\end{aligned}
\end{equation*}
At inference: encode $\mathbf{x}$, decode $\mathbf{g}$, gate experts to obtain $\mathbf{p}_{\text{pose}}$, stabilize into $\{\hat{P}_t\}$, then render video—training is end-to-end so semantic correctness and pose stability reinforce each other.

\begin{figure}[t]
    \centering
    \includegraphics[width=1\linewidth]{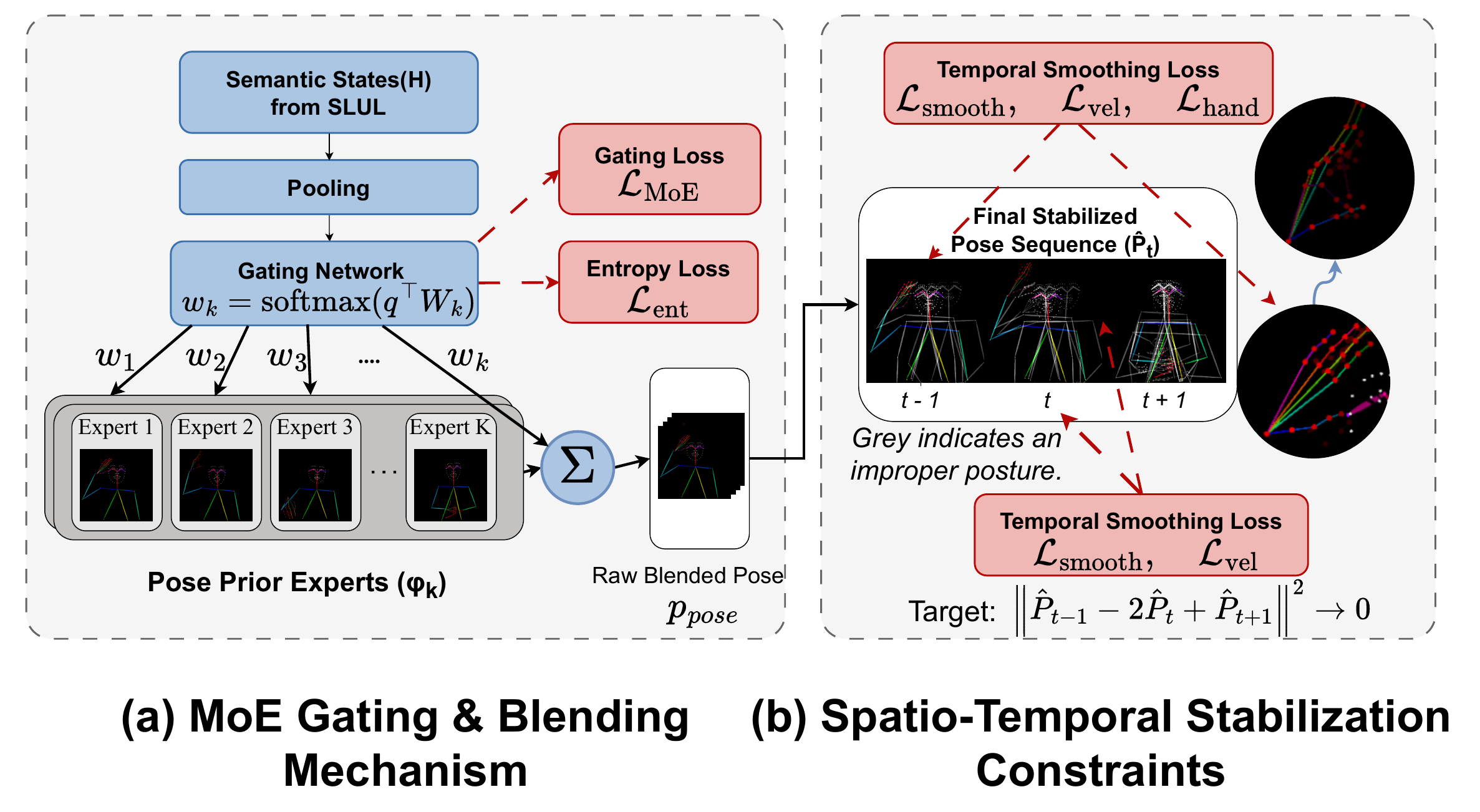}
    \vspace{-16pt}
    \caption{\textbf{Details of the SLP-MoE module:} \textcolor{purple}{\textbf{(a)}} Semantic states from SLUL generate query $q$ to produce gating weights $w_k$ over $K$ pose experts. Each expert retrieves poses from a rule-based prior database, yielding the weighted blended pose $\mathbf{p}_{\text{pose}}$. \textcolor{purple}{\textbf{(b)}} The blended poses are refined across temporal frames using smoothing loss $\mathcal{L}_{\text{smooth}}$, velocity loss $\mathcal{L}_{\text{vel}}$, and hand fidelity loss $\mathcal{L}_{\text{hand}}$ to ensure temporal coherence and spatial accuracy in the final stabilized sequence $\hat{P}_t$.}
    \label{fig:small}
    \vspace{-12pt}
\end{figure}%

%Figure~\ref{fig:small} illustrates the SLP-MoE architecture. In Figure~\ref{fig:small}(a), semantic features from SLUL generate gating weights over $K$ pose experts to produce the blended pose $\mathbf{p}_{\text{pose}}$. Figure~\ref{fig:small}(b) shows the stabilization stage where temporal smoothing, velocity, and hand fidelity losses refine the pose sequence to eliminate jitter while preserving hand accuracy. The final stabilized sequence $\{\hat{P}_t\}$ conditions our diffusion renderer for video generation.

\vspace{-6pt}
\paragraph{Pseudo-code (pose stage).} As shown in Fig. \ref{fig:small}, this stage turns gloss semantics into stable pose control signals: MoE gates specialize over pose priors, entropy keeps experts diverse, and smooth/hand/velocity penalties explicitly suppress jitter where sign intelligibility is most sensitive. The stabilized keypoints $\{\hat{P}_t\}$ then condition the diffusion renderer, so semantic correctness and motion stability are optimized jointly. Pseudo-code is given below.

\begin{algorithm}[H]
\small
\caption{SLP-MoE + stabilization}
\begin{algorithmic}[1]
\STATE $q \leftarrow \mathrm{Pool}(H)$; $w_k \leftarrow \mathrm{softmax}(q^\top W_k)$
\STATE $\mathbf{p}_{\text{pose}} \leftarrow \sum_k w_k \phi_k(\mathbf{g})$
\STATE Refine $\mathbf{p}_{\text{pose}}$ into $\{\hat{P}_t\}$, extract $\hat{H}_t$
\STATE Compute $\mathcal{L}_{\text{MoE}}$, $\mathcal{L}_{\text{ent}}$, $\mathcal{L}_{\text{smooth}}$, $\mathcal{L}_{\text{hand}}$, $\mathcal{L}_{\text{vel}}$
\STATE $\mathcal{L} \leftarrow$ combine all terms
\end{algorithmic}
\end{algorithm}
\vspace{-4pt}

Overall, our methodology is not only innovative but also addresses some of the challenges that previous work failed to notice, making our approach both robust, efficient, and achieving a very high level of performance.
\begin{table*}[t!]
\centering
\resizebox{0.99\linewidth}{!}{%
\begin{tabular}{@{}p{2.8cm}ccccc|ccccc@{}}
\toprule
 & \multicolumn{5}{c}{DEV SET} & \multicolumn{5}{c}{TEST SET} \\ 
\multicolumn{1}{c|}{Type:} & BLEU-4         & BLEU-3         & BLEU-2         & BLEU-1         & ROUGE          & BLEU-4         & BLEU-3         & BLEU-2         & BLEU-1         & ROUGE          \\ \midrule
%\multicolumn{1}{r|}{Saunders \etal \cite{saunders2020progressive}}     &  11.82 &  14.80 & 19.97 & 31.41 & 33.18 & 10.51 & 13.54 & 19.04 & 31.36 & 32.46 \\
\multicolumn{1}{r|}{NSLP-G \cite{hwang2021non}}   & - & - & - & - & - & 5.75 & 8.21 & 11.62 & 17.55 & 31.98 \\
\multicolumn{1}{r|}{Fast-SLP Transformers \cite{11099440}}  & 17.19 & 23.11 & 29.49 & 36.96 & 55.85 & 12.85 & 17.35 & 23.38 & 39.46 & 46.89 \\
%\midrule
\multicolumn{1}{r|}{Neural Sign Actors \cite{Baltatzis_2024_CVPR}}   & - & - & - & - & - & 13.12 & 18.25 & 25.44 & 41.31 & 47.55 \\
\multicolumn{1}{r|}{SignLLM-1x1B-Super-P (ASL) \cite{Fang_2025_ICCV}}   & 18.68 & 25.11 & 31.99 & 40.14 & 60.47 & 13.93 & 18.86 & 25.40 & 42.87 & 50.91 \\
\multicolumn{1}{r|}{\textbf{Stable Signer (Ours)}}   & \textbf{23.24} & \textbf{30.41} & \textbf{39.14} & \textbf{47.85} & \textbf{70.68} & \textbf{15.67} & \textbf{21.65} & \textbf{28.49} & \textbf{48.87} & \textbf{59.56} \\
\bottomrule
\end{tabular}%
}
\caption{
%Results of The First German Sign Language Production Baseline, and 
\textbf{Comparison of the Pose Video generation performance of our model:} Compared with previous models, our parameters are comparable or even fewer, and we have the burden of semantic understanding, but our performance far exceeds the latest work. ``-'' indicates that the relevant data of the work has not been made public.}
\label{tab:ourbaseline}
%\vspace{-12pt}
\end{table*}

%\textbf{Ablation Study:} 
\begin{table*}[t!]
\centering
\resizebox{0.99\linewidth}{!}{%
\begin{tabular}{@{}p{2.8cm}ccccc|ccccc@{}}
\toprule

     & \multicolumn{5}{c}{DEV SET}  & \multicolumn{5}{c}{TEST SET} \\ 
\multicolumn{1}{c|}{Approach:}  & BLEU-4         & BLEU-3         & BLEU-2         & BLEU-1         & ROUGE          & BLEU-4         & BLEU-3         & BLEU-2         & BLEU-1         & ROUGE          \\ \midrule

\multicolumn{1}{r|}{Base} & 15.09 & 21.98 & 31.07 & 59.40 & 62.09 & 12.99 & 16.07 & 26.82 & 55.32 & 58.93 \\
\multicolumn{1}{r|}{SLUL} & 22.62 & 34.23 & 48.31 & 71.16 & 72.21 & 17.09 & 22.31 & 37.13 & 58.40 & 63.22 \\
\multicolumn{1}{r|}{SLUL + SAGM Loss}& 23.24 & 30.41 & 39.14 & 47.85 & 70.68 & 15.67 & 21.65 & 28.49 & 48.87 & 59.56 \\
\multicolumn{1}{r|}{SLUL + SAGM Loss + SLP MoE}    & \textbf{25.55} & \textbf{36.79} & \textbf{47.12} & \textbf{66.79} & \textbf{78.98} & \textbf{21.03} & \textbf{24.99} & \textbf{39.03} & \textbf{57.59} & \textbf{65.26} \\
\bottomrule
\end{tabular}%
}
\caption{\textbf{Ablation Study:} A comparison of the generation performance under different settings of our model. The first three lines represent the generation of Pose Video, and the last line shows the final video generated by SLP MoE. We compared the impact of changes in different stages on the entire process, and found that our changes significantly improved the overall generation effect. \textbf{Base:} We use the base model designed by T5 \cite{2020t5}. \textbf{SLUL:} Sign Language Understanding Linker. \textbf{SLP-MoE:} Sign Language Production Mixture-of-Experts. \textbf{SAGM Loss:} Semantic-Aware Gloss Masking Loss. The results show that all the changes were beneficial.}
\label{tab:data_augmentation_results}
%\vspace{-12pt}
\end{table*}

%\newpage

\section{Experiments} \label{sec:experiments}

Here, we evaluate our Stable Signer, as well as SLUL, SLP MoE, SAGM Loss, etc. We conduct ablation evaluations, performance evaluations, efficiency evaluations, image quality evaluations, qualitative evaluations, and evaluations of prompt word meaning understanding, among others. We conduct a detailed and comprehensive test of our method. The experiments show that our method has significant improvements in all dimensions.

\subsection{Experimental Setup}

%如上文所述，我们的训练基于Prompt2Sign数据集和WLASL的美国手语部分（ASL），他们分别有30k和10k的ASL的视频片段，其中部分缺少的部分由我们手工补全或者删除。
As mentioned above, our training is based on the Prompt2Sign dataset \cite{Fang_2025_ICCV} and the American Sign Language (ASL) portion of WLASL \cite{WLASL}. They have 30k and 10k video clips of ASL respectively, and some of the missing parts (Prompts and Glosses) were properly completed or deleted.

%Our translation model utilizes the video-level translation model developed by Tarrés \etal \cite{slt-how2sign-wicv2023} and translation model \cite{camgoz2020sign}, which are refined with retraining. However, due to their model's lower performance than previous best level of mature sign language interpretation, it fails to fully showcase the accuracy of our modelf in ASL tasks. Thus, we hope that future researchers can further explore and expand in \acf{aslt}.

\subsection{Evaluation for Sign Language Production} \label{sec:ASL_experiments}

\noindent\textbf{Back Translation of ASL.}
Back translation evaluates sign language production by translating generated videos back into text and comparing with original inputs using BLEU and ROUGE metrics. 
Higher scores indicate better semantic preservation and video quality. 
We evaluate Stable Signer on the How2Sign dataset and establish baselines for ASL video production, as shown in Table \ref{tab:ourbaseline}.

Table \ref{tab:ourbaseline} shows that our method significantly outperforms existing approaches. We achieve BLEU-4 of 23.24 (dev) and 15.67 (test), substantially exceeding SignLLM's 18.68 and 13.93. Our model handles the complete pipeline including semantic understanding while achieving these improvements, making the results even more significant.

In Table \ref{tab:text_to_pose_ASL}, we present final video generation results with SLP-MoE. Compared to Fast-SLP Transformers (SignDiffusion \cite{11099440}), we achieve 48.6\% improvement on BLEU-4 (dev) and 63.7\% on test set. Our test BLEU-4 (21.03) surpasses latest Neural Sign Actors (13.12) \cite{Baltatzis_2024_CVPR} by a large margin, validating our hierarchical generation approach.

%\textbf{Back translation results}
\begin{table}[t]
\centering
\resizebox{0.99\linewidth}{!}{%
\begin{tabular}{@{}p{2.8cm}cc|cc@{}}
\toprule
     & \multicolumn{2}{c}{DEV SET}  & \multicolumn{2}{c}{TEST SET} \\ 
\multicolumn{1}{c}{Approach:}  & BLEU-4  & ROUGE & BLEU-4 & ROUGE \\ \midrule
\multicolumn{1}{r|}{Progressive Transformers \cite{saunders2020progressive}} & 15.18 & 49.46 & 15.92 & 46.57 \\ 
\multicolumn{1}{r|}{Neural Sign Actors \cite{Baltatzis_2024_CVPR}} & — & — & 13.12 & 47.55 \\ 
\multicolumn{1}{r|}{Fast-SLP Transformers \cite{11099440}} & 17.19  & 55.85 & 12.85 & 46.89 \\
\multicolumn{1}{r|}{\textbf{Stable Signer (Ours})} & \textbf{25.55}  & \textbf{78.98} & \textbf{21.03} & \textbf{65.26} \\
\multicolumn{1}{r|}{\textbf{$\Delta$ \textcolor[RGB]{34,139,34}{$Acc.$}}} & \textcolor[RGB]{34,139,34}{\textit{\textbf{+ 48.6\%}}} & \textcolor[RGB]{34,139,34}{\textit{\textbf{+ 41.4\%}}} & \textcolor[RGB]{34,139,34}{\textit{\textbf{+ 63.7\%}}} & \textcolor[RGB]{34,139,34}{\textit{\textbf{+ 39.2\%}}} \\
\bottomrule
\end{tabular}%
}
\caption{\textbf{The performance comparison results of ASL Sign Video generation:} Unlike Table \ref{tab:ourbaseline}, this is the final comparison of the effects. The 3D Avatar is regarded as a qualified sign language video. The results show that our SLP MoE will produce better effects.}
\label{tab:text_to_pose_ASL}
\vspace{-6pt}
\end{table}

%\vspace{-10pt}
\noindent\textbf{Ablation Study.}
We conduct ablation studies on the How2Sign dataset to validate each component's contribution, as shown in Table \ref{tab:data_augmentation_results}. 
We evaluate progressive improvements from base T5 \cite{2020t5} through SLUL, SAGM Loss, and SLP-MoE additions on both development and test sets.

The base model achieves BLEU-4 of 15.09 (dev) and 12.99 (test). Adding SLUL significantly improves performance, with BLEU-3 jumping from 21.98 to 34.23 on dev set, demonstrating SLUL's effectiveness in translating prompts to glosses. Incorporating SAGM Loss adjusts the metrics with ROUGE reaching 70.68, reflecting its focus on semantic-aware masking. The complete system with SLP-MoE achieves best results: BLEU-4 of 25.55 (dev) and 21.03 (test), with ROUGE scores of 78.98 and 65.26. This demonstrates that semantic integration through MoE enables effective end-to-end learning for sign language production.

\begin{figure}[t]
    \centering
    \includegraphics[width=0.99\linewidth]{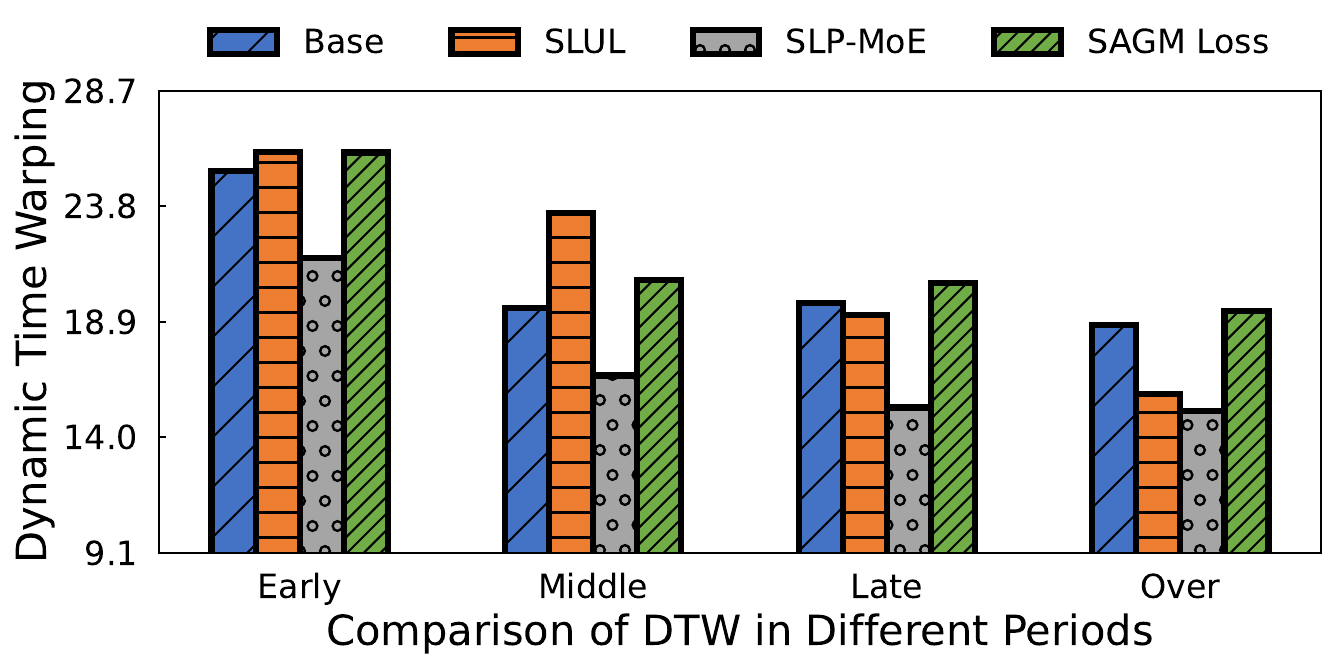}
    %\vspace{-6pt}
    \caption{\textbf{Efficiency Study:} Comparing the DTW scores of different training periods (divided by 25\% of an epoch for each training session), the lower the score, the better. We can observe that our SLUL and SLP MoE modifications have effectively improved the overall scores. The SAGM Loss, as it is a loss calculation and has no direct relation to performance, is a normal result. Therefore, we can say that all the modifications not only achieved our goals but also achieved at least a significant improvement in efficiency.}
    \label{fig:former_barline}
    \vspace{-8pt}
\end{figure}

%\vspace{-8pt}
\noindent\textbf{Training Efficiency Study.}
In Figure \ref{fig:former_barline}, we analyze training efficiency using DTW scores across epochs, partitioned into 25\% intervals. 
Lower DTW values indicate better alignment between generated and ground truth sequences.

%  is determined by epoch

Both SLUL and SLP-MoE demonstrate faster convergence and better final performance. SLUL shows accelerated learning in early stages, suggesting effective semantic understanding provides stronger supervision. SLP-MoE achieves lowest DTW scores throughout, indicating the mixture-of-experts architecture enhances both final performance and training efficiency. SAGM Loss shows comparable curves to SLUL, as expected for a loss function modification.

\begin{table}[ht]
% Add your custom packages and macros here
\newcommand{\etal}{\textit{et al.}}
%\begin{wraptable}{r}{6.0cm}
\vspace{4pt}
\centering
\resizebox{0.99\linewidth}{!}{%
\begin{tabular}{@{}p{3.0cm}ccccc@{}}
\toprule
     & \multicolumn{2}{c}{DEV SET}  & \multicolumn{2}{c}{TEST SET} \\
\multicolumn{1}{c}{Approach:}  & \multicolumn{1}{c}{BLEU-4 $\uparrow$} & \multicolumn{1}{c}{ROUGE $\uparrow$} & \multicolumn{1}{c}{BLEU-4 $\uparrow$} & \multicolumn{1}{c}{ROUGE $\uparrow$} \\ \midrule
\multicolumn{1}{r|}{Stoll \etal \cite{stoll2018sign}} & 16.34 & 48.42 & 15.26 & 48.10 \\
\multicolumn{1}{r|}{Saunders \cite{saunders2020progressive}} & 20.23 & 55.41 & 19.10 & 54.55 \\
\multicolumn{1}{r|}{Zhang \cite{10.1145/3706598.3713855}} & - & - & 27.60 & 66.40 \\
\multicolumn{1}{r|}{SignLLM \cite{Fang_2025_ICCV}} & 23.10 & 58.76 & 22.05 & 56.46 \\
\multicolumn{1}{r|}{\textbf{Stable Signer (Ours)}} & \textbf{32.08} & \textbf{76.54} & \textbf{30.74} & \textbf{69.72} \\
\multicolumn{1}{r|}{\textbf{$\Delta$ \textcolor[RGB]{34,139,34}{$Acc.$}}} & \textcolor[RGB]{34,139,34}{\textit{\textbf{+ 38.9\%}}} & \textcolor[RGB]{34,139,34}{\textit{\textbf{+ 30.2\%}}} & \textcolor[RGB]{34,139,34}{\textit{\textbf{+ 43.3\%}}} & \textcolor[RGB]{34,139,34}{\textit{\textbf{+ 23.5\%}}} \\
\bottomrule
\end{tabular}%
}
\vspace{-4pt}
\caption{\textbf{Prompt Accuracy:} Due to the scarcity of similar work to ours, we compared our semantic understanding task with the similar Text2Gloss task. Although our task was more challenging, we still outperformed previous works, as well as those in the same period.}
\label{tab:prompt}
\vspace{-5pt}
%\vspace{-14pt}
%\vspace{-0.7cm}
\end{table}

\vspace{-8pt}
\paragraph{Prompt Understanding Evaluation.} \label{paragraph:prompt_study}
We compare SLUL's Prompt2Gloss performance with previous Text2Gloss approaches in Table \ref{tab:prompt}. This is important as users naturally input prompts like \textit{``How do you sign `I carried it'?''} rather than simplified text.
Our approach achieves test BLEU-4 of 30.74 and ROUGE of 69.72, representing 43.3\% and 23.5\% improvements over SignLLM \cite{Fang_2025_ICCV}. Despite handling more complex prompts than baseline Text2Gloss tasks, we outperform all methods including \citet{10.1145/3706598.3713855} by 11.4\% on test BLEU-4. This demonstrates that SAGM Loss effectively handles prompt nuances including ambiguity and implicit semantics. Consistent performance across dev and test sets indicates robust generalization rather than overfitting.

\begin{figure*}[t]
    \centering
    \includegraphics[width=0.97\textwidth]{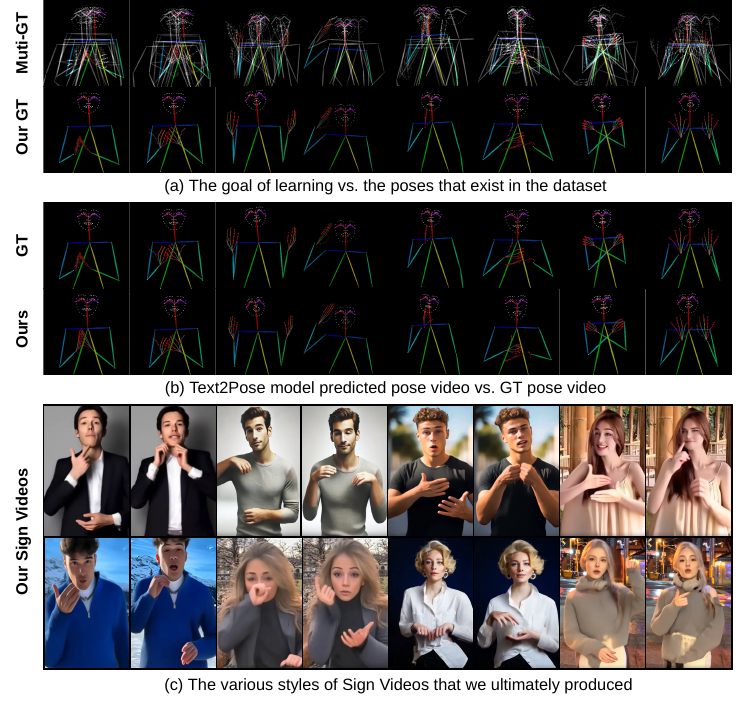}
    \vspace{-6pt}
    \caption{
    \textbf{Qualitative Results \& User Study.} We visualize our end-to-end sign language production pipeline. (a) Simplified pose targets automatically learned by the model, enabling robust end-to-end learning for SLP—an approach increasingly adopted in recent studies. (b) Intermediate pose video frames generated during the process. (c) Final sign language video frames. Ground truth frames are provided for comparison. Our method achieves high-fidelity image generation while accurately conveying sign language pose information.
    }
    \label{fig:3group}
    \vspace{-6pt}
\end{figure*}

%\vfill\break

%\input{sec/subsec/backt4GSL}

\subsection{Evaluation for Video Generation Quality} \label{sec:SignDiff_experiments}

We evaluate video generation quality through both qualitative visualizations and quantitative metrics, comprehensively assessing how our SLP-MoE module converts pose sequences into realistic, high-quality sign language videos that accurately convey semantic meaning.

%\vspace{-12pt}
\noindent\textbf{Qualitative Evaluations.} 
Figure \ref{fig:3group} provides a comprehensive visualization of our complete end-to-end sign language production pipeline across three progressive stages: (a) simplified pose representations automatically learned by the model, (b) intermediate pose video frames generated during the conversion process, and (c) final rendered sign language videos presented alongside ground truth frames for direct comparison.

The simplified pose targets shown in Figure \ref{fig:3group}(a) demonstrate a critical innovation of our approach—the model automatically learns to generate clean, semantically meaningful pose representations through the synergistic integration of rule-based pose priors and SLUL-based semantic guidance. These simplified poses effectively eliminate the noise, jitter, and temporal instability commonly present in raw pose estimations from conventional video analysis tools such as DWPose \cite{yang2023effective}. Of course, what is even more important is as shown in the first row of the picture: different sign language users have different expressions. These gestures are not aligned in time and space, which causes the models that previously learned average gestures to often be unstable, or to generate restricted or unconfident gestures. However, our method will automatically learn deterministic gestures. As long as one of the multiple gestures is completed, it is considered correct. Therefore, our generation part is more stable.

By establishing this clean pose foundation, our approach enables more robust and reliable end-to-end training—a paradigm increasingly adopted in recent sign language production research. The clean representations ensure that semantic information from SLUL is properly conveyed without being corrupted by pose estimation artifacts. This part of the assessment also indicates that our new conceptual framework has been successful.

The intermediate pose videos presented in Figure \ref{fig:3group}(b) reveal the quality and effectiveness of our Gloss2Pose conversion process. These frames showcase smooth temporal transitions between different sign gestures, accurate gesture trajectories that properly capture the dynamic nature of sign language movements, and appropriate timing that maintains the natural rhythm of signing. This approach enables the model to automatically learn the appropriate GT, making our generation process more robust and laying the foundation for the end-to-end learning of the entire complex system.
%The pose sequences demonstrate that our rule-based learnable approach successfully bridges the gap between discrete gloss tokens and continuous pose sequences while maintaining both semantic accuracy and temporal coherence. Notably, the transitions between different signs are fluid and natural, avoiding the abrupt changes or unnatural interpolations that often plague traditional pose generation methods.

Most importantly, the final generated sign language videos shown in Figure \ref{fig:3group}(c) achieve high visual fidelity and photorealistic quality that closely matches authentic human signing. When directly compared with ground truth video frames displayed alongside, our generated videos successfully maintain semantic accuracy while producing smooth, natural-looking sign language gestures.
%that would be readily intelligible to sign language users. The videos exhibit clear hand details with well-defined finger positions, natural body movements and postures, and minimal motion blur artifacts even during rapid hand movements. The diffusion-based generation approach employed in our SLP-MoE module successfully preserves fine-grained visual details such as individual finger articulations, subtle hand shape variations, and proper hand orientations—all critically important for accurate sign language comprehension. We observe minor limitations occasionally in hand orientation disambiguation during extremely rapid gesture transitions, particularly visible in the sixth column example. However, these small imperfections occur infrequently and do not significantly impact the overall semantic clarity or communicative effectiveness of the generated videos.

\begin{table}[t]
\centering
\vspace{4pt}
\resizebox{0.99\linewidth}{!}{%
\begin{tabular}{@{}p{3.0cm}cccc@{}}
\toprule
  & \multicolumn{1}{c}{SSIM $\uparrow$} & \multicolumn{1}{c}{Hand SSIM $\uparrow$} & \multicolumn{1}{c}{Hand Pose $\downarrow$} & \multicolumn{1}{c}{FID $\downarrow$} \\ \midrule
\multicolumn{1}{r|}{EDN \cite{chan2019everybody}} & 0.737 & 0.553 & 23.09 & 41.54 \\
\multicolumn{1}{r|}{vid2vid \cite{wang2018video}} & 0.750  & 0.570 & 22.51 & 56.17 \\
\multicolumn{1}{r|}{Pix2PixHD \cite{wang2018high}} & 0.737 & 0.553 & 23.06 & 42.57 \\ 
\multicolumn{1}{r|}{Stoll et al. \cite{stoll2020signsynth}} & 0.727 & 0.533 & 23.17 & 64.01 \\ 
\multicolumn{1}{r|}{\methodName{} \cite{Saunders_2022_CVPR} } & 0.759 & 0.605 & 22.05 & 27.75  \\
\multicolumn{1}{r|}{SignDiff \cite{11099440}} & 0.849 & 0.676 & 20.04 & 25.22  \\
\multicolumn{1}{r|}{Stable Signer (Ours)} & \textbf{0.892} & \textbf{0.732} & \textbf{17.68} & \textbf{21.04}  \\
\bottomrule
\end{tabular}%
}
\caption{\textbf{Video quality study:} Compared with existing and the latest hand gesture image production work, ours is the most performant, and it far exceeds the mainstream work. This is precisely why the videos we generate have better smoothness and fewer errors.}
%\caption{\textbf{Baseline model comparison} results of \ourmethodName{} sign language video generation.}
\label{tab:baselines_diverse}
%\vspace{-12pt}
\end{table}

\begin{table}[t]
\centering
%\vspace{-6pt}
\resizebox{0.99\linewidth}{!}{%
\begin{tabular}{@{}p{3.0cm}ccccc@{}}
\toprule
  & \multicolumn{1}{c}{SSIM $\uparrow$} & \multicolumn{1}{c}{Hand SSIM $\uparrow$} & \multicolumn{1}{c}{Hand Pose $\downarrow$} & \multicolumn{1}{c}{FID $\downarrow$} \\ \midrule
\multicolumn{1}{r|}{Baseline \cite{Saunders_2022_CVPR}} & 0.743 & 0.582 & 22.87 & 39.33 \\
\multicolumn{1}{r|}{Hand Discriminator} & 0.738 & 0.565 & 22.81 & 39.22 \\
\multicolumn{1}{r|}{Hand Keypoint Loss} & 0.759 & 0.605 & 22.05 & 27.75 \\
\multicolumn{1}{r|}{Zhang \cite{10.1145/3706598.3713855}} & 0.864 & - & - & 56.28  \\
\multicolumn{1}{r|}{SignDiff \cite{11099440}} & 0.849 & 0.676 & 20.04 & 25.22  \\
\multicolumn{1}{r|}{Stable Signer (No SLP-MoE) (Ours)} & \textbf{0.872} & \textbf{0.716} & \textbf{19.24} & \textbf{24.22}  \\
\multicolumn{1}{r|}{Stable Signer (Ours)} & \textbf{0.892} & \textbf{0.732} & \textbf{17.68} & \textbf{21.04}  \\
\bottomrule
\end{tabular}%
}
\caption{\textbf{Ablation Study of SLP-MoE's Video Quality:} We also attempted to conduct an ablation study on our SLP-MoE to verify whether it was the improvement of the base model or our method that was responsible for the effect. The results showed that although the better backbone played a role, our method could achieve even greater improvements.}
%\vspace{-4pt}
%\caption{\textbf{Ablation study results:} comparison of our \FrameReinforcementNet{} and others can enhance image quality \cite{Saunders_2022_CVPR}.}
\label{tab:ablation_diff}
% \vspace{}
\vspace{-12pt}
\end{table}

\noindent\textbf{Baseline Comparison.} 
We conduct comprehensive quantitative comparisons with SOTA sign language video generation methods using four complementary evaluation metrics (SSIM\&Hand SSIM \cite{wang2004image}, Hand Pose error measuring \cite{ge20193d}, FID \cite{heusel2017gans}) that capture different aspects of generation quality.
%(1) Structural Similarity Index (SSIM) computed over the entire video frame to measure overall visual similarity and structural preservation , (2) Hand SSIM calculated specifically on cropped hand regions to assess the quality of the most semantically critical parts of sign language videos, (3) Hand Pose error measuring the absolute pixel distance between predicted and ground truth 2D hand keypoints using a pre-trained pose estimation model , and (4) Fréchet Inception Distance (FID) for evaluating perceptual quality and photorealism of generated images . These metrics collectively provide a comprehensive assessment of both geometric accuracy and perceptual quality.
Table \ref{tab:baselines_diverse} shows Stable Signer achieves superior performance across all metrics, our image quality is also far superior to that of SOTA or the latest works. 
%We attain SSIM of 0.892, significantly exceeding SignDiff (0.849) and \methodName{} (0.759). Hand SSIM of 0.732 indicates excellent hand detail preservation crucial for intelligibility. Hand Pose error of 17.68 shows 11.8\% improvement over SignDiff (20.04), critical as small position errors can alter meaning. Our FID of 21.04 represents best perceptual quality. These improvements stem from: (1) end-to-end training optimizing semantic accuracy and visual quality jointly, (2) SLP-MoE effectively integrating semantic information, and (3) rule-based pose priors reducing noise for cleaner generation.

\begin{figure}[t]
    \centering
    \includegraphics[width=0.99\linewidth]{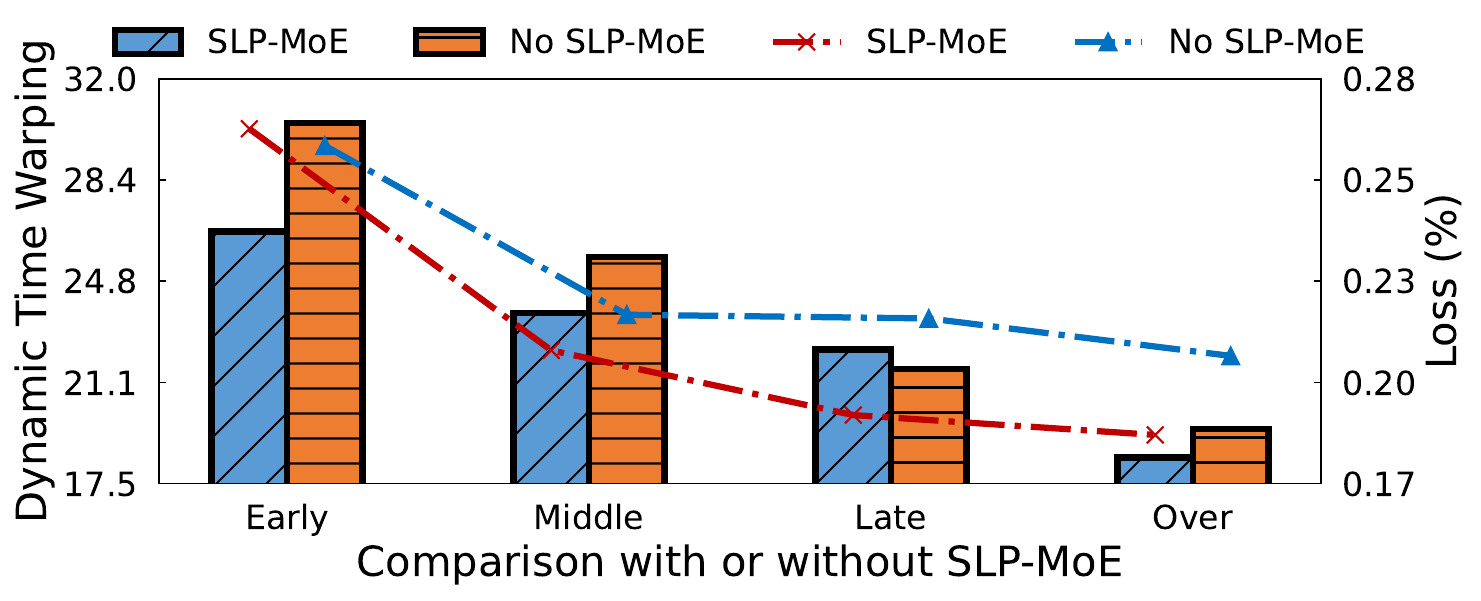}
    %\vspace{-6pt}
    \caption{\textbf{Ablation Study on SLP-MoE's Training Efficiency:} We examined the efficiency under the condition where there was no SLP-MoE (using the regular Pose2Vid model instead \cite{cheng2025wan}). The results showed that using our SLP-MoE for joint training could achieve more thorough convergence, train faster, and produce better results.}
    \label{fig:diff_barline}
    \vspace{-12pt}
\end{figure}%
%\vspace{-12pt}

\noindent\textbf{Ablation Study on Video Generation.}
To isolate SLP-MoE's contribution and verify improvements come from methodology rather than just better backbones, we conduct ablations in Table \ref{tab:ablation_diff}.

Results show clear progressive improvements. 
%From baseline (0.743 SSIM), individual enhancements like Hand Discriminator provide modest gains. Recent work (Zhang, SignDiff) shows strong performance with SignDiff achieving 0.849 SSIM. 
Our ``No SLP-MoE'' variant achieves 0.872 SSIM and 24.22 FID, outperforming SignDiff and validating that our simplified pipeline and SLUL already provide benefits. Full Stable Signer with SLP-MoE reaches 0.892 SSIM and 21.04 FID, demonstrating MoE adds significant value. Hand Pose improves from 19.24 to 17.68 (8.1\%), showing semantic integration helps generate more accurate positions critical for intelligibility.

Figure \ref{fig:diff_barline} shows SLP-MoE achieves better final performance and faster convergence. DTW scores decrease more rapidly with MoE, suggesting semantic information provides stronger learning signals. The performance gap remains substantial at convergence, indicating MoE enables better optimization. Loss curves show comparable convergence rates, confirming gains come from enhanced capability to leverage semantics rather than easier optimization. This validates our hypothesis that end-to-end joint training of semantic understanding and generation yields superior results.

\section{Conclusion} \label{sec:conclusion}
In this paper, we observed that the current goals of the sign language generation tasks have fallen into suboptimal average goals due to multiple Pose Ground Truth. Therefore, we attempted to shorten the production chain and developed the Stable Signer framework for end-to-end SLP, which includes SLUL for complex semantic understanding, SLP-MoE for generating high-quality sign language videos, and SAGM Loss for end-to-end training. The results show that, in the ASL production process, with the same dataset, it can achieve a significant improvement of approximately 50\% in the accuracy of poses compared to the existing traditional methods. Moreover, it has achieved results that surpass the state-of-the-art in nearly ten comprehensive and rich experiments.

\section*{Limitations}

%我们在任务不同姿势的衔接上会稍微有点闪烁，但是我们的SLP MoEs会消除这个问题，根据定性评估，这不影响最终效果。
There might be a slight flickering in the transition between different poses of the task, but our SLP MoEs will eliminate this problem. According to the qualitative assessment, this does not affect the final outcome.
%

% Bibliography entries for the entire Anthology, followed by custom entries
%\bibliography{custom,anthology-overleaf-1,anthology-overleaf-2}

% Custom bibliography entries only
{
    \small
    \bibliography{ref/main, ref/sds, ref/meta, ref/llm, ref/rl, ref/slt, ref/how2sign}
}

\appendix

\newpage

\section{Related Works}
\label{sec:related}

\subsection{Sign Language Production}
%手语生成（Sign Language Production, SLP）旨在从口语输入生成手语视频或动作序列，弥合聋人和听障群体的沟通障碍。早期的深度学习方法将SLP分解为级联步骤：文本到词缀（T2G）、词缀到姿态（G2P）和姿态到手语（P2S）。然而，这些流水线方法在手语动作之间的平滑过渡方面存在困难。近期研究探索了使用Transformer架构和扩散模型的端到端方法。Neural Sign Actors提出了一种基于扩散的模型，在4D手语虚拟形象上训练，使用基于SMPL-X骨架的解剖学图神经网络来实现逼真的3D手语生成。T2S-GPT引入了动态向量量化用于自回归手语生成，而SignLLM则通过新颖的强化学习组件将SLP扩展为支持八种手语的多语言框架。尽管取得了这些进展，生成具有精细手部和面部表情的逼真、语义准确的手语动作仍然具有挑战性。
Sign Language Production (SLP) aims to generate sign language videos or motion sequences from spoken language inputs, bridging communication gaps for the deaf and hard-of-hearing community \cite{gregory2013bridgingcommunicationgapswiththedeaf}. Early deep learning approaches decomposed SLP into cascaded steps: Text-to-Gloss (T2G) \cite{abdullah2025stateofthearttranslationtexttoglossusing}, Gloss-to-Pose (G2P) \cite{xie2023g2pddmgeneratingsignpose}, and Pose-to-Sign (P2S) \cite{moryossef2025posebasedsignlanguageappearance}. However, these pipeline methods struggled with smooth motion blending between signs. Recent works have explored end-to-end approaches using transformer architectures and diffusion models. Neural Sign Actors \cite{baltatzis2024neuralsignactorsdiffusion} proposed a diffusion-based model trained on 4D signing avatars with anatomically informed graph neural networks defined on the SMPL-X skeleton for realistic 3D sign production. T2S-GPT \cite{yin2024t2sgptdynamicvectorquantization} introduced dynamic vector quantization for autoregressive SLP, while SignLLM \cite{Fang_2025_ICCV} extended SLP to a multilingual framework supporting eight sign languages through novel reinforcement learning components. Despite these advances, generating realistic, semantically accurate signing motions with fine-grained hand and facial expressions remains challenging \cite{walsh2025slrtp2025signlanguageproduction,TII-Web-2023-Falcon,Cho-arxiv-2022-Prompt,fang2025signxfoundationmodelsign,hwang2021non}.
%in sign language.

%Subsequently, researchers began to introduce human pose estimation technology \cite{Katz-IEEE-1987-estimation,Ko-2019-SLT-based-human-keypoint-estimation,charles2014automatic,SLT,SLT_pose_amit,yin2022mlslt}, converting sign language videos into skeleton sequences for recognition, which significantly improved the models' ability to perceive action details \cite{saunders2020progressive, saunders2021continuous}. With the advancement of computer vision technology, various powerful pose estimation methods \cite{cai2023smplerx,yang2023effective} have emerged. However, because of their different data formats and representation methods, finding a way to effectively integrate this heterogeneous information has become an urgent challenge \cite{stoll2020text2sign, huang2021towards,fang2025signllmsignlanguageproduction,fang2025signdiffdiffusionmodelamerican}. Notably, lexical units (signs) play an important role in SL understanding. As shown in Table \ref{tab:SLT_datasets}, different datasets focus on different output goals. Understanding also requires taking account of the complex syntax of these languages. This involves not only linear sign order, but also interactions with non-manual signals, which play an essential grammatical role in signed languages. Gloss-based representations like ``I-GO-STORE'' can help, in part, to model the unique linguistic rules of a given signed language \cite{camgoz2018neural, camgoz2020sign,gloss,WordsAreOurGlosses,zelinka2020neural,neidle2022alternative,zhou-etal-2024-multimodal,neidle2022aslvideocorpora}. 

\subsection{Motion Generation}
%随着扩散模型的采用，人体运动生成取得了显著进展。MotionDiffuse是最早的基于扩散的文本驱动运动生成框架之一，通过去噪步骤展示了概率映射，并实现了细粒度的身体部位控制。MDM（Motion Diffusion Model）引入了基于Transformer的架构和无分类器引导，预测运动样本而非噪声，以便于对关节位置和速度施加几何损失。LGTM提出了一种局部到全局的流水线，利用大语言模型将运动描述分解为特定身体部位的叙述，解决了局部语义准确性问题。近期工作还探索了音乐到舞蹈的合成、组合动作生成和风格化运动控制。这些进展为生成时间连贯的人体运动提供了基础技术，可直接应用于手语运动合成，其中精确的手部关节运动和身体协调至关重要。
Human motion generation has witnessed significant progress with the adoption of diffusion models. MotionDiffuse \cite{zhang2022motiondiffusetextdrivenhumanmotion} was among the first diffusion-based frameworks for text-driven motion generation, demonstrating probabilistic mapping through denoising steps and enabling fine-grained body part control. MDM (Motion Diffusion Model) \cite{tevet2022humanmotiondiffusionmodel} introduced a transformer-based architecture with classifier-free guidance, predicting motion samples rather than noise to facilitate geometric losses on joint positions and velocities. LGTM \cite{Sun_2024} proposed a Local-to-Global pipeline leveraging large language models to decompose motion descriptions into part-specific narratives, addressing local semantic accuracy issues. Recent works have also explored music-to-dance synthesis, compositional action generation \cite{liu2024languagefreecompositionalactiongeneration}, and stylized motion control \cite{zhong2025smoogptstylizedmotiongeneration}. These advances provide foundational techniques for generating temporally coherent human movements, which are directly applicable to sign language motion synthesis where precise hand articulation and body coordination are essential.

%Beyond traditional pose estimation, there are also emerging technologies. PrimeDepth \cite{zavadski2024primedepth}, specifically designed for depth estimation, can provide 3D structural information about scenes \cite{qiu2021dense, carreira-2017-i3d}. Sapiens segmentation \cite{khirodkar2024sapiens} provides a new perspective for interpreting signers' fine-grained actions through precise human body part segmentation \cite{guler2018densepose, saunders2021mixed}. These advances provide rich feature representation methods for SL recognition, offering good extensions to the information dimensions of pose estimation \cite{7298594, koller2020quantitative}.

\begin{figure}[t]
    \centering
    \includegraphics[width=0.65\linewidth]{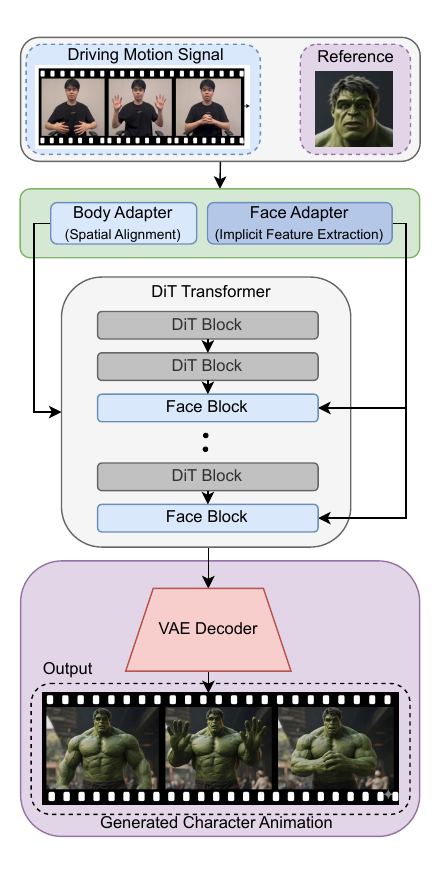} 
    \caption{The framework of Wan-Animate \cite{cheng2025wan}. It utilizes a unified input paradigm and a decoupled control strategy. Spatially-aligned skeleton signals are directly injected into the DiT backbone for body control, while facial expressions are driven by implicit features extracted from reference images and injected via cross-attention layers.}
    \label{fig:wan_animate}
\end{figure}

\subsection{Controllable Video Generation}
%可控视频生成已成为一个活跃的研究方向，将基于图像的控制机制扩展到时间域。ControlNet开创性地通过零卷积层将深度图、边缘检测和人体姿态等结构控制整合到文本到图像的扩散模型中。ControlVideo在无需微调的情况下将ControlNet适配到视频生成，通过完全跨帧交互和交错帧平滑继承了高质量一致的帧生成。ControlNeXt通过将可训练参数减少高达90%进一步提高了效率，同时支持包括视频扩散姿态序列在内的多种条件控制。Follow-Your-Pose和MimicMotion专门针对带有姿态引导的人体视频生成。这些可控生成技术为手语视频合成提供了有前景的方向，其中基于姿态条件的生成可以确保解剖学正确和语义对齐的手语动作。
Controllable video generation has emerged as an active research direction, extending image-based control mechanisms to the temporal domain. ControlNet \cite{zhang2023addingconditionalcontroltexttoimage} pioneered the integration of structural controls such as depth maps, edge detection, and human pose into text-to-image diffusion models through zero convolution layers. ControlVideo \cite{zhang2023controlvideotrainingfreecontrollabletexttovideo} adapted ControlNet to video generation without fine-tuning, inheriting high-quality consistent frame generation through fully cross-frame interaction and interleaved-frame smoothing. ControlNeXt \cite{peng2025controlnextpowerfulefficientcontrol} further improved efficiency by reducing trainable parameters up to 90\% while supporting diverse conditional controls including pose sequences for video diffusion. Follow-Your-Pose \cite{ma2024followposeposeguidedtexttovideo} and MimicMotion \cite{zhang2025mimicmotionhighqualityhumanmotion} specifically addressed human video generation with pose guidance. 
% --- 新增 Wan-Animate 内容 ---
Most recently, Wan-Animate \cite{cheng2025wan} advances this field by leveraging a Diffusion Transformer (DiT) \cite{peebles2023scalablediffusionmodelstransformers,tevet2022humanmotiondiffusionmodel,fang2025streamflowtheoryalgorithmimplementation,NEURIPS2022_ec795aea} architecture. As illustrated in Figure \ref{fig:wan_animate}, it employs a decoupled control paradigm that integrates spatially-aligned skeleton signals for body movements and implicit facial features via cross-attention for fine-grained expression reenactment.
% ---------------------------
These controllable generation techniques offer promising directions for sign language video synthesis \cite{SLcomprehensibility,signor,Xu-SIGPLAN-2022-Systematic}, where pose-conditioned generation can ensure anatomically correct and semantically aligned signing motions.

%The success of diffusion models lies in their unique noise addition and denoising process, which enables models to learn the inherent structure of data. Through gradual denoising, models can generate high-quality samples \cite{wang2018video, stoll2020text2sign}. 
%This characteristic 
%It makes diffusion models particularly suitable for handling feature conversion problems in SL recognition, especially when processing continuously changing pose sequences \cite{duarte2021how2sign, camgoz2018neural}.
%\input{sec/subsec/preliminary}
\section{Discussion}

% 在这篇文章中，我们进行了充分而完备的实验：对于各种ASL生成效果的实验，我们在公平条件下，对比了目前的最新的手语工作，并且我们比他们有非常巨大的提高。对于消融评估，由于各种组件有先后依赖，我们逐步添加验证我们的改动都是有效的。对于效率评估和效率指标的消融评估，我们都进行了全方面的实验和对比实验。定性评估和图像质量评估都明显展现出我们模型质量的优越和方法的优势。
In this paper, we conducted thorough and comprehensive experiments: for various ASL generation effects, we compared the current latest sign language work under fair conditions, and we achieved a significant improvement over them. For ablation assessment, due to the sequential dependencies of various components, we gradually added to verify that our modifications were effective (all components, including the a priori component, were evaluated). For efficiency assessment and efficiency metric ablation assessment, we conducted comprehensive experiments and comparative experiments. Qualitative assessment and image quality assessment clearly demonstrated the superiority of our model quality and the advantages of our method.

%不过，我们也注意到最近有些生成工作比如SignCLIP \cite{jiang-etal-2024-signclip}、SignAlignLLM \cite{inan-etal-2025-signalignlm}，Sign as Token \cite{zuo2025signs}, etc工作，他们与我们的目标相似但是不同，而且使用不同的数据集训练，因此我们不需要也无法轻易比较。截至目前，我们应该是性能最好的ASL SLP工作：很多手语工作由于轻微的目标不同，导致是不能随便说我们没有和某些工作比较的。
But, we also noticed that some recent generation works such as SignCLIP \cite{jiang-etal-2024-signclip}, SignAlignLLM \cite{inan-etal-2025-signalignlm}, Sign as Token \cite{zuo2025signs}, etc., share similar goals with ours but are different. Moreover, they use different datasets for training. Therefore, we cannot easily compare them. Up to now, our work on ASL SLP should be the best: many sign language works cannot be directly compared with ours due to minor differences in goals. Finally, the non-ASL production is beyond our scope of work. We can consider conducting research on this in the future.

\end{document}